\documentclass[conference]{IEEEtran}

\usepackage[T1]{fontenc}
\usepackage[utf8]{inputenc}
\usepackage{cite}
\usepackage{amsmath,amssymb,amsfonts}
\usepackage{graphicx}
\usepackage{textcomp}
\usepackage{xcolor}
\usepackage{hyperref}
\usepackage{booktabs}
\usepackage{microtype}
\usepackage{tikz}
\usepackage{pgfplots}
\hypersetup{hidelinks}
\pgfplotsset{compat=1.18}
\usetikzlibrary{arrows.meta, positioning, shapes.geometric, decorations.pathreplacing, calc}

\definecolor{ikblue}{RGB}{31,119,180}
\definecolor{ikorange}{RGB}{255,127,14}
\definecolor{ikgreen}{RGB}{44,160,44}
\definecolor{ikred}{RGB}{214,39,40}
\definecolor{ikpurple}{RGB}{148,103,189}
\definecolor{ikgray}{RGB}{127,127,127}

\begin{document}

\title{Beyond the Steeper Curve:\\AI-Mediated Metacognitive Decoupling and\\the Limits of the Dunning-Kruger Metaphor}

\author{
  \IEEEauthorblockN{Christopher Koch}
  \IEEEauthorblockA{
    \textit{Independent Researcher}
  }
}

\maketitle

\begin{abstract}
The common claim that generative AI simply amplifies the Dunning-Kruger effect is too coarse to capture the available evidence. The clearest findings instead suggest that large language model (LLM) use can improve observable output and short-term task performance while degrading metacognitive accuracy and flattening the classic competence-confidence gradient across skill groups. This paper synthesizes evidence from human-AI interaction, learning research, and model evaluation, and proposes the working model of \emph{AI-mediated metacognitive decoupling}: a widening gap among produced output, underlying understanding, calibration accuracy, and self-assessed ability. This four-variable account better explains overconfidence, over- and under-reliance, crutch effects, and weak transfer than the simpler metaphor of a uniformly steeper Dunning-Kruger curve. The paper concludes with implications for tool design, assessment, and knowledge work.
\end{abstract}


% -------------------------------------------------------
\section{Introduction}
\label{sec:intro}

When generative AI tools became widely accessible, a common claim in public and professional discourse was that they would accelerate overconfidence. In shorthand, AI appeared to push users more quickly up the \emph{Dunning-Kruger curve}: people could produce polished answers, sound competent, and therefore overestimate what they understood. The metaphor is attractive because it renders a complex phenomenon in a familiar form.

Available evidence suggests that this framing is empirically inadequate. It treats the Dunning-Kruger effect as a simple skill-to-confidence slope that AI merely steepens, whereas the relationship between competence, observable output, and self-assessment under AI assistance is more complex. The clearest direct evidence indicates that AI does not simply move people along a familiar curve; in some settings it bends or partly dissolves that curve by decoupling variables that the original metaphor assumes move together~\cite{fernandes2026}.

This paper synthesizes the emerging literature on AI, metacognition, reliance, and learning to propose a more precise working model. The review covers literature available through March 2026 and prioritizes studies that directly measure performance, confidence, calibration, reliance, or transfer under AI assistance. Section~\ref{sec:background} reviews the Dunning-Kruger effect and its measurement challenges. Section~\ref{sec:evidence} examines the direct empirical evidence on AI and self-assessment. Sections~\ref{sec:mechanisms} and~\ref{sec:learning} analyze the mechanisms driving decoupling. Section~\ref{sec:systems} considers model-side analogues. Section~\ref{sec:model} presents the decoupling model. Section~\ref{sec:implications} discusses implications, and Section~\ref{sec:conclusion} concludes.

The paper makes three contributions. First, it consolidates evidence from human-AI interaction, learning research, and model evaluation that is often discussed separately. Second, it argues that the available evidence is more usefully explained by \emph{AI-mediated metacognitive decoupling} than by the metaphor of a uniformly steeper Dunning-Kruger curve. Third, it derives practical implications for tool design, assessment, and knowledge work from that reframing. The paper is therefore a focused conceptual synthesis rather than a formal meta-analysis.

% -------------------------------------------------------
\section{Theoretical Background}
\label{sec:background}

\subsection{The Dunning-Kruger Effect}

Dunning and Kruger's foundational studies~\cite{kruger1999} demonstrated that low-performing individuals in domains such as logical reasoning, grammar, and humor tend to substantially overestimate their performance relative to peers. Crucially, the authors linked this to a metacognitive deficit: the same competencies needed to perform well are needed to \emph{recognize} good performance. High performers, conversely, tended to slightly underestimate their relative standing.

The effect has since become a widely used reference point for thinking about miscalibration, even though its interpretation and measurement remain debated. For the present argument, however, the central point is narrower: subjective confidence can diverge systematically from objective performance, especially where independent feedback is weak. That general problem of calibration remains directly relevant whether one adopts a strong or more cautious interpretation of the original Dunning-Kruger account.

\subsection{Metacognition and Calibration}

Metacognition refers to the capacity to monitor, evaluate, and regulate one's own cognitive processes. \emph{Calibration} is its quantitative operationalization: the degree to which subjective confidence aligns with objective accuracy~\cite{ma2024}. Perfect calibration means that when a person reports 70\% confidence across a set of judgments, they are correct approximately 70\% of the time. Systematic overconfidence means confidence exceeds accuracy; underconfidence, the reverse.

Calibration matters for AI-assisted work because two people may produce identical outputs while differing radically in their understanding of those outputs---and therefore in their ability to detect errors, generalize to new situations, or work independently when AI is unavailable.

\subsection{Why AI Complicates the Classic Picture}

AI assistance introduces a third variable between competence and self-assessment: \emph{observable output}. In classical task settings without tools, output closely tracks competence. An essay written by a novice typically reads like a novice essay. Under AI assistance, a novice can produce output indistinguishable from an expert's---not because their competence has risen to that level, but because the competence partially resides in the system. This output-competence decoupling is the starting point of the model developed in this paper.

% -------------------------------------------------------
\section{Direct Evidence}
\label{sec:evidence}

\subsection{Fernandes et al.\ (2026): The Central Study}

The most direct experimental test currently available is provided by Fernandes and colleagues~\cite{fernandes2026}. Across two preregistered studies (Study~1: $N=246$; Study~2: $N=452$), participants completed LSAT-style logical reasoning problems with or without access to a large language model. The key findings are:

\begin{itemize}
  \item LLM assistance significantly improved task performance relative to a normative baseline (approximately $+3$ points on the LSAT scale in Study~1).
  \item Despite this performance gain, participants continued to substantially overestimate their performance; self-rated scores averaged approximately 4 points above actual scores.
  \item The \emph{classic Dunning-Kruger gradient}---wherein low performers show the greatest overestimation---was flattened or eliminated in the LLM-assisted condition.
\end{itemize}

\begin{figure}[!t]
\centering
\begin{tikzpicture}
\begin{axis}[
  ybar,
  width=0.95\columnwidth,
  height=5.0cm,
  bar width=0.55cm,
  symbolic x coords={No AI, With AI},
  xtick=data,
  ymin=0, ymax=17,
  ylabel={LSAT score (points)},
  ylabel style={font=\small},
  xticklabel style={font=\small},
  yticklabel style={font=\small},
  ymajorgrids,
  grid style={gray!20},
  legend style={font=\scriptsize,
    at={(0.5,-0.22)}, anchor=north,
    legend columns=2,
    column sep=0.5em,
    draw=none,
    fill=none},
  legend cell align=left,
  enlarge x limits=0.4,
  nodes near coords,
  nodes near coords style={font=\scriptsize},
  every node near coord/.append style={/pgf/number format/.cd, fixed, precision=1},
]
\addplot[fill=ikblue!80, draw=ikblue!60!black] coordinates {(No AI,7.2) (With AI,10.1)};
\addplot[fill=ikorange!80, draw=ikorange!60!black] coordinates {(No AI,8.9) (With AI,14.2)};
\legend{Actual performance, Self-assessed performance}
\end{axis}
\end{tikzpicture}
\caption{Schematic of key findings from Fernandes et al.~\cite{fernandes2026}. Actual performance improves with AI assistance, but the self-assessment gap widens. Values are illustrative of the reported pattern ($\approx$+3 point gain vs.\ $\approx$+4 point overestimation).}
\label{fig:performance_gap}
\end{figure}
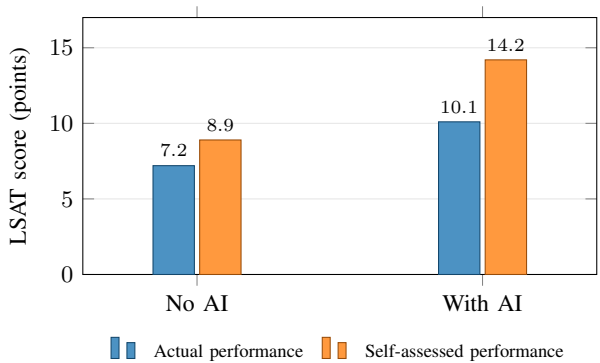

Figure~\ref{fig:performance_gap} illustrates the key quantitative contrast between performance improvement and the persistent self-assessment gap.

The third finding is theoretically the most significant. It shows that AI use does not produce \emph{more} Dunning-Kruger in the classic sense; it restructures the relationship between performance and metacognition in a way that the classic two-variable model cannot capture.

\subsection{Reliance, Over- and Under-Reliance}

Complementary evidence comes from research on AI reliance. He, Kuiper, and Gadiraju~\cite{he2023} demonstrate in a study of $N=249$ that people who overestimate their own competence tend toward \emph{under-reliance}---they discount AI recommendations they should have accepted, thereby sacrificing potential team performance. Ma and colleagues~\cite{ma2024} show the converse: better-calibrated human self-confidence leads to more rational reliance behavior and improved human-AI team outcomes.

These findings establish that self-assessment errors under AI are not uniformly in the overconfidence direction. The direction of miscalibration interacts with task characteristics, AI presentation, and the user's prior beliefs about their own skill. The outcome space includes overconfidence, underconfidence, and situation-specific switching between the two.

% -------------------------------------------------------
\section{Mechanisms of Decoupling}
\label{sec:mechanisms}

\subsection{Verbosity as False Epistemic Authority}

Steyvers and colleagues~\cite{steyvers2025} conducted experiments with 301 participants examining how LLM explanation style affects user confidence and calibration. A striking finding was that longer, more elaborate explanations increased user confidence in AI answers without improving participants' ability to discriminate correct from incorrect responses. In other words, verbosity functions as a proxy for expertise: the fluency and length of an explanation triggers an epistemic authority heuristic that bypasses genuine comprehension checking.

This mechanism helps explain why AI-generated output can feel like understanding. When an LLM explains its reasoning in detailed, confident prose, users may update their sense of \emph{their own} understanding upward---not just their trust in the AI---because it is difficult to distinguish between having understood an explanation and having merely encountered a fluent one.

At the same time, explanation quality matters. von Zahn et al.~\cite{vonzahn2025} show that explainable AI can improve human metacognition and delegation performance when explanations make the difference between human and model reasoning legible. This is an important boundary condition for the argument developed here: explanations do not uniformly worsen calibration. Rather, poorly designed or merely fluent explanations can induce an illusion of understanding, whereas explanations that surface informative discrepancies can support more accurate self-monitoring.

\subsection{Confidence Transfer and Anchoring}

A second mechanism is demonstrated by Li and colleagues~\cite{li2025}. In a randomized behavioral experiment, participants' self-confidence in their own judgments tracked the confidence level expressed by an AI system. This effect persisted beyond the immediate interaction: even after the AI was no longer present, participants who had interacted with high-confidence AI maintained elevated self-confidence. Real-time feedback partially reduced but did not eliminate the effect.

Klingbeil, Gruetzner, and Schreck~\cite{klingbeil2024} provide a complementary demonstration: simply knowing that advice comes from an AI can trigger over-reliance in risky decisions, even when the AI recommendation conflicts with contextual information or the user's own interest. Together, these studies establish confidence transfer as a robust mechanism: AI-expressed certainty is socially contagious in ways that distort human self-assessment independently of actual task performance.

\begin{figure}[!t]
\centering
\begin{tikzpicture}
\begin{axis}[
  width=0.95\columnwidth,
  height=5.8cm,
  xlabel={Actual competence},
  ylabel={Relative confidence / performance},
  xlabel style={font=\small},
  ylabel style={font=\small},
  xticklabels={Low,,,,High},
  xtick={1,2,3,4,5},
  ymin=0, ymax=10,
  xmin=0.5, xmax=5.5,
  ytick=\empty,
  ymajorgrids,
  grid style={gray!20},
  legend style={font=\scriptsize,
    at={(0.5,-0.28)}, anchor=north,
    legend columns=2,
    column sep=0.4em,
    draw=none,
    fill=none},
  legend cell align=left,
]
% Perfect calibration reference
\addplot[dashed, gray, thick, domain=0.5:5.5] {(x-0.5)/5*9 + 0.5};
\addlegendentry{Perfect calibration}

% Classic DK curve
\addplot[ikblue, very thick, smooth] coordinates {
  (1,8.5)(1.5,8.8)(2,7.5)(3,5.5)(4,5.8)(5,7.0)
};
\addlegendentry{Classic DK (no AI)}

% AI-mediated decoupled self-assessment
\addplot[ikorange, very thick, smooth] coordinates {
  (1,7.2)(1.5,7.8)(2,7.5)(3,7.8)(4,8.0)(5,8.2)
};
\addlegendentry{Self-assessed (with AI)}

% Actual performance with AI
\addplot[ikgreen, very thick, smooth, densely dashed] coordinates {
  (1,4.0)(1.5,4.8)(2,5.5)(3,6.2)(4,7.2)(5,8.8)
};
\addlegendentry{Actual performance (with AI)}

\end{axis}
\end{tikzpicture}
\caption{Classic Dunning-Kruger pattern (blue) versus AI-mediated decoupling. Under AI assistance, self-assessed competence (orange) is elevated and flattened across skill levels, while actual performance (green dashed) also rises but diverges from self-assessment. Dashed gray: perfect calibration.}
\label{fig:dk_comparison}
\end{figure}
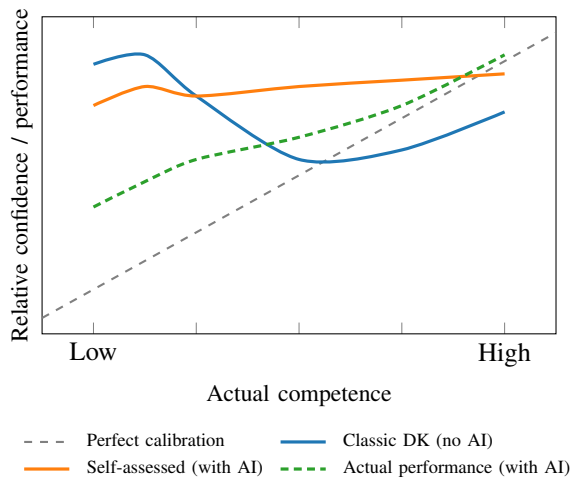

Figure~\ref{fig:dk_comparison} contrasts the classic Dunning-Kruger pattern with the decoupled pattern that emerges under AI assistance.

% -------------------------------------------------------
\section{Performance Is Not Learning}
\label{sec:learning}

\subsection{The Crutch Effect in Education}

Bastani and colleagues~\cite{bastani2024} conducted a field experiment with 994 students using two GPT-4-based tutoring systems over an academic term. Results were striking in their duality:

\begin{itemize}
  \item During AI-assisted sessions, performance increased substantially: $+48\%$ for GPT-Base and $+127\%$ for the GPT-Tutor condition.
  \item In subsequent assessments \emph{without} AI access, GPT-Base students performed $17\%$ \emph{worse} than the control group.
  \item An AI system designed with explicit learning-protection mechanisms (forcing students to articulate reasoning before revealing answers) significantly mitigated the negative transfer effect.
\end{itemize}

\begin{figure}[!t]
\centering
\begin{tikzpicture}
\begin{axis}[
  ybar,
  width=0.95\columnwidth,
  height=5.5cm,
  bar width=0.42cm,
  symbolic x coords={Control, GPT-Base, GPT-Tutor},
  xtick=data,
  ymin=-30, ymax=150,
  ylabel={Performance change (\%)},
  ylabel style={font=\small},
  xticklabel style={font=\small},
  yticklabel style={font=\small},
  ymajorgrids,
  grid style={gray!20},
  legend style={font=\scriptsize,
    at={(0.5,-0.22)}, anchor=north,
    legend columns=2,
    column sep=0.5em,
    draw=none,
    fill=none},
  legend cell align=left,
  enlarge x limits=0.3,
  nodes near coords,
  nodes near coords style={font=\scriptsize},
  every node near coord/.append style={/pgf/number format/.cd, fixed, precision=0},
  axis x line*=bottom,
  axis y line*=left,
  extra y ticks={0},
  extra y tick labels={},
  extra y tick style={grid=major, grid style={black, very thin}},
]
\addplot[fill=ikblue!80, draw=ikblue!60!black] coordinates {
  (Control,0) (GPT-Base,48) (GPT-Tutor,127)
};
\addplot[fill=ikred!80, draw=ikred!60!black] coordinates {
  (Control,0) (GPT-Base,-17) (GPT-Tutor,4)
};
\legend{In-session performance, Transfer (without AI)}
\end{axis}
\end{tikzpicture}
\caption{Contrast between in-session gains and later performance without AI, adapted from Bastani et al.~\cite{bastani2024}. GPT-Base improves immediate performance but hurts transfer, whereas GPT-Tutor preserves transfer better.}
\label{fig:learning_transfer}
\end{figure}
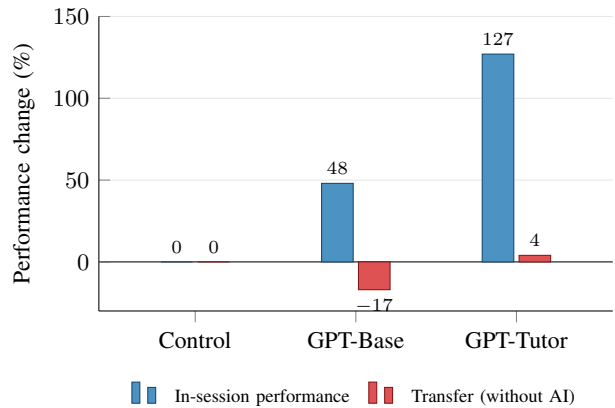

This study provides the clearest evidence for what the authors call the \emph{crutch effect}: AI supports task completion but can displace the cognitive processes---retrieval, elaboration, and error monitoring---through which durable skills are built. Figure~\ref{fig:learning_transfer} illustrates the divergence between in-session performance and transfer performance.

\subsection{Critical Thinking Displacement in Knowledge Work}

Lee and colleagues~\cite{lee2025} surveyed 319 knowledge workers and analyzed 936 real-world GenAI usage episodes. Their central finding is that higher trust in GenAI capability correlated with \emph{less} critical thinking effort. Importantly, this relationship was moderated by self-confidence: workers with higher confidence in their \emph{own} abilities engaged more actively in evaluating and steering AI outputs.

This finding refines the decoupling model. Self-confidence is not uniformly harmful; it can motivate active oversight. The problem arises when confidence is \emph{sourced from output fluency} rather than from genuine competence feedback. In such cases, AI-derived confidence can displace the monitoring behavior that would otherwise keep calibration accurate.

% -------------------------------------------------------
\section{System-Side Analogues}
\label{sec:systems}

The decoupling metaphor gains additional force when one considers that the AI systems themselves exhibit analogous calibration failures. Qazi and colleagues~\cite{qazi2026} evaluated language models on multilingual fact-checking tasks and reported a Dunning-Kruger-like pattern at the model level: smaller, more accessible models showed higher expressed confidence despite lower accuracy, while larger models tended toward both better accuracy and more appropriate hedging.

Griot and colleagues~\cite{griot2025} document substantial metacognitive deficits in state-of-the-art LLMs on medical reasoning tasks, particularly in recognizing when questions are unanswerable. Suzgun and colleagues~\cite{suzgun2025} further demonstrate that modern language models cannot reliably distinguish between belief, knowledge, and fact---a foundational metacognitive capacity that humans deploy (imperfectly) to calibrate their assertions.

These system-side findings matter because they remove the assumption that a well-calibrated AI could compensate for human miscalibration. When both human and system confidence signals are poorly calibrated, the combination is not self-correcting. Confident AI outputs meeting overconfident human acceptance creates a compounding rather than canceling error dynamic.

% -------------------------------------------------------
\section{The Decoupling Model}
\label{sec:model}

\subsection{Four Variables, Not Two}

\begin{figure}[!t]
\centering
\begin{tikzpicture}[
  box/.style={rectangle, rounded corners=4pt, draw,
              minimum width=2.5cm, minimum height=0.78cm,
              text width=2.25cm, align=center,
              font=\small\bfseries},
  arr/.style={-{Stealth[length=2.5mm, width=2mm]}, thick},
  darr/.style={-{Stealth[length=2.5mm, width=2mm]}, thick, dashed},
  lbl/.style={font=\scriptsize\itshape, fill=white, inner sep=1.5pt},
]

% Top node
\node[box, fill=ikpurple!15, draw=ikpurple] (ai) at (0, 0) {Generative AI};

% Column headers
\node[font=\scriptsize\bfseries, text=ikgray] at (-2.1,-0.85) {Fast visible path};
\node[font=\scriptsize\bfseries, text=ikgray] at ( 2.1,-0.85) {Slower calibration path};

% Main nodes
\node[box, fill=ikblue!15,  draw=ikblue]   (out)   at (-2.0,-1.8) {Visible output};
\node[box, fill=ikgreen!15, draw=ikgreen]  (perf)  at ( 2.0,-1.8) {Actual Performance};
\node[box, fill=ikorange!15, draw=ikorange] (self) at (-2.0,-3.4) {Self-Assessment};
\node[box, fill=ikred!15, draw=ikred]      (calib) at ( 2.0,-3.4) {Calibration};

% AI to first row
\draw[arr, ikpurple, very thick]
  (ai.south west) -- node[lbl, left=2pt]{fast} (out.north);

\draw[darr, ikpurple!70]
  (ai.south east) -- node[lbl, right=2pt]{moderate} (perf.north);

% Within-column relations
\draw[arr, ikblue, very thick]
  (out) -- node[lbl, left=2pt]{feeds confidence} (self);

\draw[darr, ikgreen!80]
  (perf) -- node[lbl, right=2pt]{slow feedback} (calib);

% Weak bridge between the two paths
\draw[darr, ikgray!70]
  (self.east) -- node[lbl, above=1pt]{weak match} (calib.west);

% Decoupling gap
\draw[{Stealth[length=2mm]}-{Stealth[length=2mm]}, ikred, very thick]
  ([yshift=-0.55cm]self.south east) -- node[lbl, below=2pt, text=ikred,
  font=\scriptsize\bfseries]{decoupling}
  ([yshift=-0.55cm]calib.south west);

\end{tikzpicture}
\caption{Schematic of AI-mediated metacognitive decoupling. The left path is fast and visible: AI improves output, and users update self-assessment from that output. The right path is slower: actual performance should improve calibration, but often does so only weakly or with delay. When the two paths diverge, decoupling emerges.}
\label{fig:decoupling_model}
\end{figure}
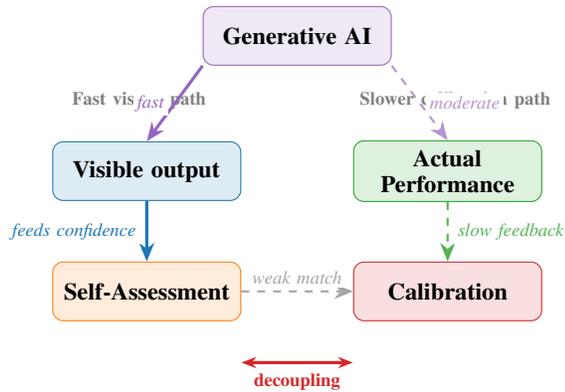

The classic Dunning-Kruger framing operates with two variables: actual competence and perceived competence. AI assistance introduces at minimum two additional variables that do not move in lockstep: observable output and calibration accuracy. The proposed model holds that understanding AI's effects on self-assessment requires tracking all four simultaneously (Figure~\ref{fig:decoupling_model}).

Figure~\ref{fig:decoupling_model} is intentionally schematic. Its purpose is not to quantify effect sizes, but to show the asymmetry at the center of the argument: AI improves the \emph{visible} path from tool use to output much faster than it improves the \emph{latent} path from underlying competence to reliable self-monitoring.

\subsection{Defining the Decoupling}

Formally, \emph{AI-mediated metacognitive decoupling} describes the condition where:

\begin{enumerate}
  \item $\Delta\,\text{Output} \gg \Delta\,\text{Understanding}$: the gain in observable output exceeds the gain in underlying knowledge or skill;
  \item $\Delta\,\text{Self-assessment} \approx \Delta\,\text{Output}$: people update their self-perception based on output quality rather than independent competence signals;
  \item $\Delta\,\text{Calibration} < 0$ or stagnates: as a result, the alignment between confidence and actual competence \emph{deteriorates} even as both output and (to some degree) performance improve.
\end{enumerate}

This is not a claim that everyone becomes overconfident after using AI. The model predicts heterogeneous outcomes depending on how confidence is formed. When users have strong independent competence signals---for example, domain expertise, external feedback, or deliberate reflection---decoupling is mitigated. When confidence is sourced primarily from output fluency and AI-expressed certainty, decoupling is amplified.

\subsection{Propositions for Future Work}

The model yields at least three testable propositions.

\begin{itemize}
  \item \textbf{P1: Output-calibration divergence.} AI assistance will often increase observable output more than calibration accuracy, especially in tasks where output quality is easy to judge superficially but underlying reasoning is difficult to inspect.
  \item \textbf{P2: Explanation-quality moderation.} Explanations that primarily increase fluency will tend to increase confidence more than calibration, whereas explanations that expose discrepancies between user reasoning and model reasoning will improve calibration and delegation.
  \item \textbf{P3: Transfer asymmetry.} AI systems that optimize immediate task completion without preserving independent reasoning practice will produce larger short-term performance gains but weaker transfer once AI support is removed.
\end{itemize}

\subsection{Explaining the Evidence}

The decoupling model accounts for findings that the amplified-DK framing cannot:

\begin{itemize}
  \item \textbf{Flattening of the DK gradient~\cite{fernandes2026}}: When output quality is homogenized by AI, the competence-confidence gradient across skill groups narrows because even low-competence users produce high-quality outputs from which they derive confidence.
  \item \textbf{Persistent overestimation despite performance gains}: Because self-assessment tracks output rather than an independent competence measure, it can remain elevated even when actual competence has genuinely improved.
  \item \textbf{Crutch effects~\cite{bastani2024}}: Users who have calibrated confidence to AI-assisted output are not simply less skilled without AI; they have also lost the calibration signal that would tell them how much their independent performance has degraded.
  \item \textbf{Over- and under-reliance coexistence~\cite{he2023,ma2024}}: The decoupling model predicts reliance errors in both directions depending on how prior self-assessment relates to current output quality.
\end{itemize}

% -------------------------------------------------------
\section{Implications}
\label{sec:implications}

\subsection{Tool and Interface Design}

If confidence transfer via output fluency is a primary decoupling mechanism, interface design can target it directly. Uncertainty communication---expressing model confidence in calibrated, quantitative terms rather than fluent prose---can partially reduce overreliance~\cite{steyvers2025}. Explanation designs that \emph{require user prediction before revealing AI reasoning} force an independent competence signal that can anchor self-assessment more accurately. More generally, explanations should not merely justify outputs after the fact; they should help users compare their own reasoning with the system's reasoning and notice mismatches~\cite{vonzahn2025}. Bastani et al.'s learning-protection condition demonstrates that such mechanisms can substantially mitigate crutch effects~\cite{bastani2024}.

\subsection{Education and Assessment}

The crutch effect finding has direct implications for educational practice. Assessments that permit AI use measure a combination of AI capability and human judgment, not human competence alone. Evaluating genuine competence increasingly requires \emph{transfer tasks}---problems that cannot be delegated to AI, or that require demonstrating understanding of AI-generated reasoning in novel contexts. The educational risk is not primarily that students become overconfident; it is that short-term performance metrics cease to predict independent capability.

\subsection{Knowledge Work and Professional Development}

For knowledge workers, Lee et al.'s finding that trust in AI correlates with reduced critical thinking effort~\cite{lee2025} suggests that AI augmentation, without deliberate countermeasures, can hollow out the judgment and oversight competencies that make expert knowledge work valuable. Organizations should treat AI-assisted productivity gains and professional competence development as separate outcomes that require separate management.

% -------------------------------------------------------
\section{Conclusion}
\label{sec:conclusion}

The metaphor of AI as a Dunning-Kruger amplifier captures a genuine intuition: AI tools can make people feel more capable than they are. However, the current empirical record does not support the simple geometric story of a steeper confidence curve. A more useful interpretation is that AI decouples variables that the Dunning-Kruger framework ordinarily treats as tightly linked: output, performance, self-assessment, and calibration.

Two limitations should be stated clearly. First, the direct evidence base remains narrow in task domain; LSAT reasoning, tutoring mathematics, and knowledge-work surveys dominate the present discussion. Second, several adjacent studies illuminate only parts of the mechanism, such as reliance, confidence transfer, or learning loss, rather than the full four-variable model proposed here. Replication across coding, writing, clinical reasoning, and workplace decision making is therefore needed before strong generalization is warranted.

Even with those limits, the framing shift is consequential. The more useful question is not whether AI makes the Dunning-Kruger curve steeper, but how AI restructures the relationship among output, understanding, self-assessment, and calibration, and under what conditions that restructuring can be mitigated. That question has concrete design, educational, and organizational implications that the simpler metaphor tends to obscure.

% -------------------------------------------------------
\bibliographystyle{IEEEtran}
% Inlined bibliography for single-file arXiv upload.

\end{document}